\documentclass[10pt,twocolumn,letterpaper]{article}

\usepackage{cvpr}              

\usepackage{graphicx}
\usepackage{amsmath}
\usepackage{amssymb}
\usepackage{booktabs}

\usepackage[pagebackref,breaklinks,colorlinks]{hyperref}

\usepackage[capitalize]{cleveref}
\crefname{section}{Sec.}{Secs.}
\Crefname{section}{Section}{Sections}
\Crefname{table}{Table}{Tables}
\crefname{table}{Tab.}{Tabs.}

\def\cvprPaperID{6754} 
\def\confName{CVPR}
\def\confYear{2022}

\begin{document}

\title{Transferable End-to-end Room Layout Estimation via Implicit Encoding}

\author{Hao Zhao$^{1,2}$, Rene Ranftl$^2$, Yurong Chen$^2$, Hongbin Zha$^1$ \\
	$^1$Peking University $^2$Intel Labs\\
	{\tt\small \{zhao-hao@,zha@cis.\}pku.edu.cn}\\
	{\tt\small \{hao.zhao,rene.ranftl,yurong.chen\}@intel.com}}
\maketitle

\begin{abstract}
   We study the problem of estimating room layouts from a single panorama image. Most former works have two stages: feature extraction and parametric model fitting. Here we propose an end-to-end method that directly predicts parametric layouts from an input panorama image. It exploits an implicit encoding procedure that embeds parametric layouts into a latent space. Then learning a mapping from images to this latent space makes end-to-end room layout estimation possible. However end-to-end methods have several notorious drawbacks despite many intriguing properties. A widely raised criticism is that they are troubled with dataset bias and do not transfer to unfamiliar domains. Our study echos this common belief. To this end, we propose to use semantic boundary prediction maps as an intermediate domain. It brings significant performance boost on four benchmarks (Structured3D, PanoContext, S3DIS, and Matterport3D), notably in the zero-shot transfer setting. Code, data, and models will be released.
\end{abstract}


\section{Introduction}

Room layout estimation is the task of recovering parametric room structure elements (e.g., walls, floors and ceilings) from images. If efficient and effective room layout estimation is finally achieved in the future, many robotics and graphics applications could benefit from it. We notice that existing state-of-the-art methods all share a common paradigm of \emph{fitting on features}, which consists of two stages. In the first stage, a fully convolutional neural network extracts semantic cues from inputs. Semantic cues may come in various forms like keypoints, boundaries or facets. In the second stage, parametric representations are fitted on these cues, using tailored cost functions. 

Different from these methods, we pursue end-to-end room layout estimation. There are works \cite{chen2020bsp}\cite{li2019supervised} that predict compact parametric shape representations for objects from clean inputs. Yet whether it is possible for cluttered scenes remains unclear. We believe giving a positive answer to this question is methodologically meaningful. The other motivation is building a solution that naturally enjoys the ongoing progress of deep learning. Now that the problem is addressed by an end-to-end neural network, new generic techniques can be incorporated seamlessly, including but not limited to layers, losses or training schemes. 


Specifically, we resort to the idea of implicit encoding \cite{park2019deepsdf}\cite{mescheder2019occupancy}\cite{chen2020bsp}. A shape is represented as a latent vector, on which an implicit function is conditioned. The latent vector lives in a space that is a surrogate of the structured output space of room layouts, on which we can build discriminative models. This space bridges the gap between sensory inputs and parametric representations, making end-to-end room layout estimation possible. 

However, end-to-end methods often struggle with generalization performance. When training such a naive end-to-end model on the largest known synthetic dataset, Structured3D \cite{zheng2020structured3d}, we achieves $81.93\%$ top-view IoU. Unfortunately, evaluating the trained model on other three small-scale benchmarks, PanoContext \cite{zhang2014panocontext}, S3DIS \cite{armeni2016s3dis} and Matterport3D \cite{zou2021manhattan}, leads to $67.04\%$, $51.43\%$ and $24.68\%$ IoUs, respectively. This echoes the common belief that end-to-end methods are troubled with dataset bias \cite{agrawal2018assume}\cite{zhao2019lds}. Even worse, fine-tuning on these smaller datasets often yields even lower performance. We delve into this problem and identify two sources of domain drift. 

The first is related to a biased shape embedding regressor. Datasets vary in both low-level and high-level properties. Photo-realistic rendering in Structured3D cannot reproduce all subtle visual effects caused by real-world material and lighting, and consequently differs in low-level details to real images. S3DIS pre-dominantly features office rooms while other datasets do not. Office rooms do not cover typical home furniture, making it different from other datasets in term of high-level scene composition. As such, the first domain drift results from regressing layout shape embedding from RGB inputs of different statistics. The second is related to a biased shape embedding space. Similar to the widely studied domain-adaptive road scene parsing problem \cite{richter2016playing}, the second source can be considered comparable to the issue of label set mismatch.


We develop two techniques to address domain drift:

The first is to pre-process panoramic images into semantic boundary prediction maps (Fig.~\ref{fig:fcn}). It is inspired by semantic transfer \cite{zhao2017physics}, but differs from it in terms of motivation. \cite{zhao2017physics} intends for proper initialization, while our aim is to address zero-shot transferability. This technique works surprisingly well, despite the fact that this map is spatially sparse (i.e., most pixels have nearly zero values) and suffers from radial distortion. Specifically, it improves the performance on PanoContext, S3DIS and Matterport3D to $84.43\%(+17.39\%)$ , $82.69\%(+31.26\%)$ and $73.73\%(+49.05\%)$, respectively. Notably, this performance boost is achieved without fine-tuning.

The second is to improve the implicit encoding step with an enormous amount of synthetic data. Using synthetic data for 3D scene understanding is an actively explored topic, yet has seen limited success thus far \cite{su2015render}\cite{song2017ssc}\cite{zhang2017physically}. Our domain, top-view layout occupancy image, is very simple and hardly influenced by sim-to-real gap. Translating a wall in the occupancy image faithfully generates in-domain new samples. Specifically, we use the training set of Structured3D \cite{zheng2020structured3d} as anchors and generate one million synthetic samples via a conditional uniform augmentation strategy. It leads to nearly perfect implicit self-encoding performance on PanoContext ($98.91\%$) and S3DIS ($98.88\%$). Again, this is achieved without fine-tuning.



\section{Related Work}

\textbf{Room Layout Estimation.} The task of estimating room layouts from perspective images was first introduced by \cite{hedau2009recovering}. Line segments are clustered according to three Manhattan directions \cite{coughlan1999manhattan}. Sampling lines originating from three orthogonal vanishing points yields room layout proposals, which can be later ranked by a discriminative model \cite{tsochantaridis2005large}. Handcrafted statistical features like geometric context \cite{hoiem2005popup} or orientation maps \cite{lee2009geometric} show strong discriminative power for this task. An interesting feature of this scheme is that 3D object boxes can be sampled and ranked in a similar way, naturally leading to the joint parsing of objects and layouts \cite{hedau2010thinking}\cite{lee2010estimating}. A further extension is to build a Bayesian model that is aware of the prior distribution of object-layout relationship. Although inference is usually expensive, better 3D scene parsing results can be achieved \cite{zhao2013scene}\cite{choi2015gp}.

After the advent of deep learning, robust features generated by fully convolutional architectures improved room layout estimation performance by large margins. Pixel-wise semantic features come in various forms like boundaries \cite{mallya2015learning}\cite{zhao2017physics}, facets \cite{dasgupta2016delay}\cite{ren2016cfile} or keypoints \cite{lee2017roomnet}\cite{huang2018hopr}. However, these methods still rely on post-processing algorithms to generate parametric layout results. These algorithms can be as simple as keypoint linking \cite{lee2017roomnet} or as complex as MCMC sampling \cite{huang2018hopr}. We call this paradigm \emph{fitting on features}.

Panoramic room layout estimation was first proposed by \cite{zhang2014panocontext} which adapts perspective techniques to this problem. Panoramic images have a $360^\circ$ field of view, thus are suitable for inferring the layout of a whole room. LayoutNet \cite{zou2021manhattan} learns keypoint/boundary cues from panoramic RGB and remapped segment images, using deep networks. HorizonNet \cite{sun2019horizonnet} introduces the tailored column-wise representation, making use of calibrated panoramic images whose rows are exactly parallel to horizons. Dula-Net \cite{yang2019dula} learns deep occupancy cues on the remapped top view of panoramic images. CFL \cite{fernandez2020cfl} explores the usage of equirectangular convolution for better representation learning on distorted panoramic images. However, all these methods are designed based on the \emph{fitting on features} scheme. Different from them, we propose the first end-to-end room layout estimation method that directly predicts parametric layouts without fitting in a post-processing stage.

\textbf{Parametric Reconstruction.} Morphable models are principled representations for parametric reconstruction. It has been successfully used for faces \cite{blanz1999morphable}, body \cite{loper2015smpl} and chairs \cite{wu2018interpreter}. High-dimensional geometric data is projected onto a set of base shapes, and the coefficient vector is treated as a compact parametric representation. As such, regressing these coefficients naturally allows end-to-end parametric reconstruction from sensory inputs, e.g., using deep neural networks. However, it is not clear if these techniques can be applied to the shape space of room layouts. Deforming a rectangular room into a non-convex room with 14 walls is different from deforming a neutral face to a smiling one. To this end, we borrow the recently proposed idea of deep binary space partitioning \cite{chen2020bsp}. We choose a set of hyperplanes as the parameterization, which can be converted to an implicit representation for differentiable self supervision. 

\textbf{Synthetic Data.} Using synthetic data for deep learning is an exciting topic as it virtually provides an unlimited amount of data. But the photo-realism of current rendering techniques is still far from satisfactory. Render4CNN \cite{su2015render} shows improved pose estimation performance using cropped object images. But this success does not transfer to pixel-wise detailed understanding as evidenced by the limited accuracy boost reported in \cite{zhang2017physically}. We argue that the right way to use synthetic data is to choose a proper domain that is less influenced by rendering artifacts. For example, semantic scene completion \cite{song2017ssc}\cite{zhang2018sgc} benefits from an enormous corpus of synthetic depth images because rendering only geometry is obviously easier than rendering images.

\section{Method}


\begin{figure*}[ht]
	\includegraphics[width=0.75\linewidth]{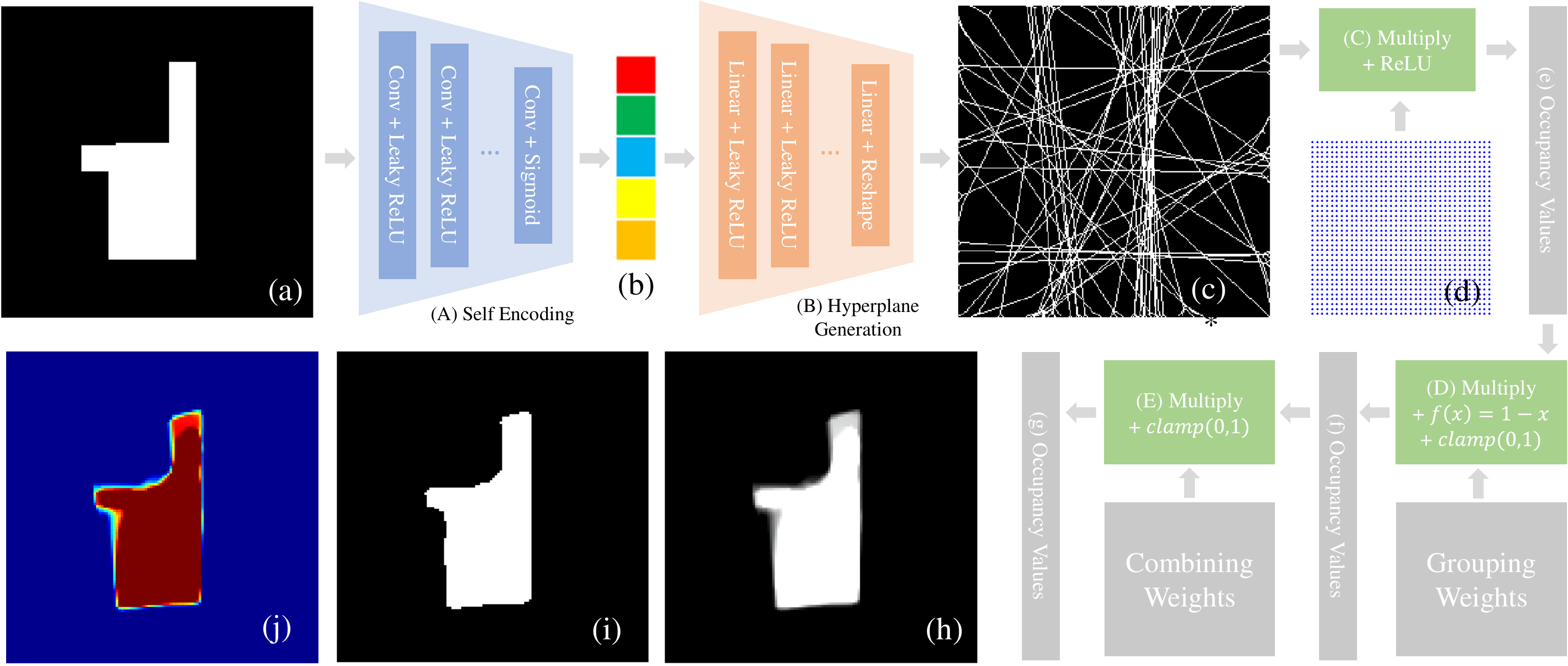}
	\centering
	\caption{
		An illustration of implicit encoding. (h) can be supervised by accessing input (a) with coordinates in (d). (i) is generated by thresholding (h) with 0.5. (j) is the same as (h) but visualized with color temperature. Other details are described in text.
	}
	\label{fig:encoding}
\end{figure*}

\subsection{Implicit Encoding}

The objective of implicit encoding is to turn a room layout into a latent code, so that we can predict the code using a neural network. Getting a code during test time means getting its corresponding layout, which is exactly what we pursue: end-to-end room layout estimation. We choose the (rasterized) top-down occupancy image (Fig.~\ref{fig:encoding}-a) as the representation of the room layout. Training an auto-encoder on this image with a pixel-wise reconstruction loss is a natural way to get the code. However, in this way, the code can only be used to recover a rasterized occupancy image, which is of limited use in many scenarios. 

Instead of pixel-wise auto-encoding, our implicit encoding module (detailed in Fig.~\ref{fig:encoding}) firstly maps a layout into a code, which can be later mapped to a set of hyper-planes. Intriguingly, the parameters of hyper-planes are trained in a self-supervised manner, i.e., we don't provide ground truth parameters for training. This is achieved through an implicit occupancy function that links the hyper-plane parameters with the input image. Specifically speaking, occupancy values can be calculated with the hyper-plane parameters and supervised by accessing corresponding positions in the input occupancy image, as formally stated below.

\subsubsection{Self-encoding and Hyper-plane Generation}

The input occupancy image (Fig.~\ref{fig:encoding}-a) is denoted as $O_i\in \mathbb{R}^{R\times R}$, which is generated by projecting the layout to the top-down viewpoint. The representation of $O_i$ is binary, with in-room pixels having value $1$ and out-of-room pixels having value $0$. The resolution $R$ is a hyper-parameter. Figure.~\ref{fig:encoding}-A shows the architecture of a self-encoding convolutional network, whose final activation function is Sigmoid. This Sigmoid function keeps the latent code (Fig.~\ref{fig:encoding}-b) magnitude-constrained, which is critical to the success of regressing the code in the subsequent step. The self-encoding network is formally denoted as $\pi_\phi^{\rm SE}$:

\begin{equation}
	\label{eq:se}
	\pi_\phi^{\rm SE}:\mathbb{R}^{R\times R} \rightarrow \mathbb{R}^{D}
\end{equation}

The $D$-dimensional latent code $C_i \in \mathbb{R}^{D}$ is generated by $C_i=\pi_\phi^{\rm SE}(O_i)$. The latent code is is a compact representation of the room layout yet there is no way to understand the meaning of its each component. We map $C_i$ into a set of oriented hyper-planes, each of which partitions the space into two parts. One side of the hyper-plane is occupied by the room while the other is not. This mapping is done with a hyper-plane generation network denoted by $\pi_\psi^{\rm HG}$:

\begin{equation}
	\label{eq:se}
	\pi_\psi^{\rm HG}:\mathbb{R}^{D} \rightarrow \mathbb{R}^{N_p\times 3}
\end{equation}

$\pi_\psi^{\rm HG}$ is implemented by a multi-layer perceptron which outputs another array so that the final layer reshapes it into the size of $N_p\times 3$. The set of hyper-planes $P_i\in \mathbb{R}^{N_p\times 3}$ is generated by $P_i=\pi_\psi^{\rm HG}(C_i)$. Here $N_p$ is a hyper-parameter representing the number of hyper-planes. $N_p$ is set to a fixed value much larger than the number of walls in practice. Each hyper-plane has three parameters corresponding to the coefficients in $ax+by+c=0$. A visualization of $P_i$ is given in Fig.~\ref{fig:encoding}-c. The coordinate system of $ax+by+c=0$ is set such that the origin is at the center of the input image $O_i$ and two axes align with the pixel coordinates.

As a reminder, our learning system is self-supervised so we do not have the ground truth hyper-plane equations for $P_i$. In fact, it is not even possible to annotate $N_p$ hyper-plane equations for each layout. In order to enforce $P_i$ to naturally have the meaning we want, we render $P_i$ back to input image $O_i$, in a differentiable manner. 

\subsubsection{Differentiable Rendering}

This differentiable rendering is achieved through an implicit occupancy function, with the help of a set of continuous coordinates in the coordinate system of $ax+by+c=0$. These coordinates only serve as a surrogate in our training, so that they neither receive gradients nor undergo optimization. 

These coordinates are illustrated as Fig.~\ref{fig:encoding}-d. In practice, we use homogeneous representation so that coordinates are denoted as $c\in \mathbb{R}^{3\times N_c}$. As shown in Fig.~\ref{fig:encoding}-C, we multiply $P_i$ with $c$ and apply a $\rm ReLU$ function, getting the initial occupancy values $o^1={\rm ReLU}(P_ic)$. $o^1\in \mathbb{R}^{N_p\times N_c}$ corresponds to Fig.~\ref{fig:encoding}-e. The reason why $o^1$ is \emph{initial} is that we still cannot impose supervision on it. We can get $N_c$ ground truth occupancy values by bilinearly accessing $O_i$. However, it is not yet possible to get $N_p\times N_c$ ground truth occupancy values to supervise $o^1$. 

The first step to convert $N_p\times N_c$ values to $N_c$ values is grouping. We resort to a set of learnalbe grouping weights $W_g\in \mathbb{R}^{N_s\times N_p}$. The intuition is that this step would group hyper-planes to $N_s$ shape primitives. We first multiply $W_g$ with $o^1$, then apply a $f(x)=1-x$ function to align with the training data definition and finally impose a clamp function $g(x)={\rm max}({\rm min}(x,0),1)$ (as illustrated in Fig.~\ref{fig:encoding}-D). As such we can get the intermediate occcpancy value $o^2=g(f(W_g o^1))\in \mathbb{R}^{N_s\times N_c}$, shown as Fig.~\ref{fig:encoding}-f. 

The final step is to combine $N_s$ shape primitives into a final layout shape. This is achieved by another set of learnable combining weights $W_c\in \mathbb{R}^{1\times N_s}$. As shown in Fig.~\ref{fig:encoding}-E, we multiply $W_c$ with $o^2$ and clamp it. The final occupancy value is generated by $o^3=g(W_c o^2)\in \mathbb{R}^{N_c}$, which corresponds to Fig.~\ref{fig:encoding}-g. 

Now that all components in Fig.~\ref{fig:encoding} has been described, we can write them together in a single equation:

\begin{equation}
	\label{eq:ie}
	o^3=g(W_cg(f(W_g{\rm ReLU}(\pi_\psi^{\rm HG}(\pi_\phi^{\rm SE}(O_i))c))))
\end{equation}

The original input is the rasterized occupancy image $O_i$ and the homogeneous representation of continuous coordinates $c$. The output is the occupancy values $o^3$ at coordinates $c$. The ground truth occupancy values $\overline{o}_i=O_i(c)$ can be obtained by accessing coordinates $c$ bilinearly in $O_i$. There are four sets of network parameters in this equation: $\phi$, $\psi$, $W_g$ and $W_c$. These parameters can be trained by forcing the final occupancy $o^3$ towards $\overline{o}_i$. Different from conventional self-encoding, the input is a rasterized representation while the output is an implicit representation, although they correspond to the same underlying signal.

\subsubsection{Loss Functions}

Formally, the first loss is occupancy value reconstruction: 

\begin{equation}
	\label{eq:l1}
	L_o=\sum_{i}\|o^3_i-\overline{o}_i\|^2
\end{equation}

We need regularization terms to ensure the behaviors of $W_g$ and $W_c$. Following \cite{chen2020bsp}, we enforce each element in $W_g$ to be bounded in $[0,1]$ so that it behaves like a soft grouping operator. The loss term is implemented as such:

\begin{equation}
	\label{eq:l2}
	L_g=\sum_{t\in W_g}{\rm max}(t-1,0)-\sum_{t\in W_g}{\rm max}(t,0)
\end{equation}

Finally, we drag the sum of $W_c$ towards 1 so that it functions as a combining operator:

\begin{equation}
	\label{eq:l3}
	L_c=\sum_{t\in W_c}|t-1|
\end{equation}

These three loss functions altogether supervise the implicit encoding network. Note that they correspond to occupancy values, grouping weights and combining weights, all at the scale of $[0,1]$. So we add them using the same balancing coefficient. 

\begin{equation}
	\label{eq:lse}
	\arg\min_{\phi,\psi,W_g,W_c}L_o+L_g+L_c
\end{equation}

This implicit encoding can effectively reconstruct complicated rooms like the one shown in Fig.~\ref{fig:encoding}. 

\subsubsection{Data Augmentation}

\begin{figure}[ht]
	\includegraphics[width=0.8\linewidth]{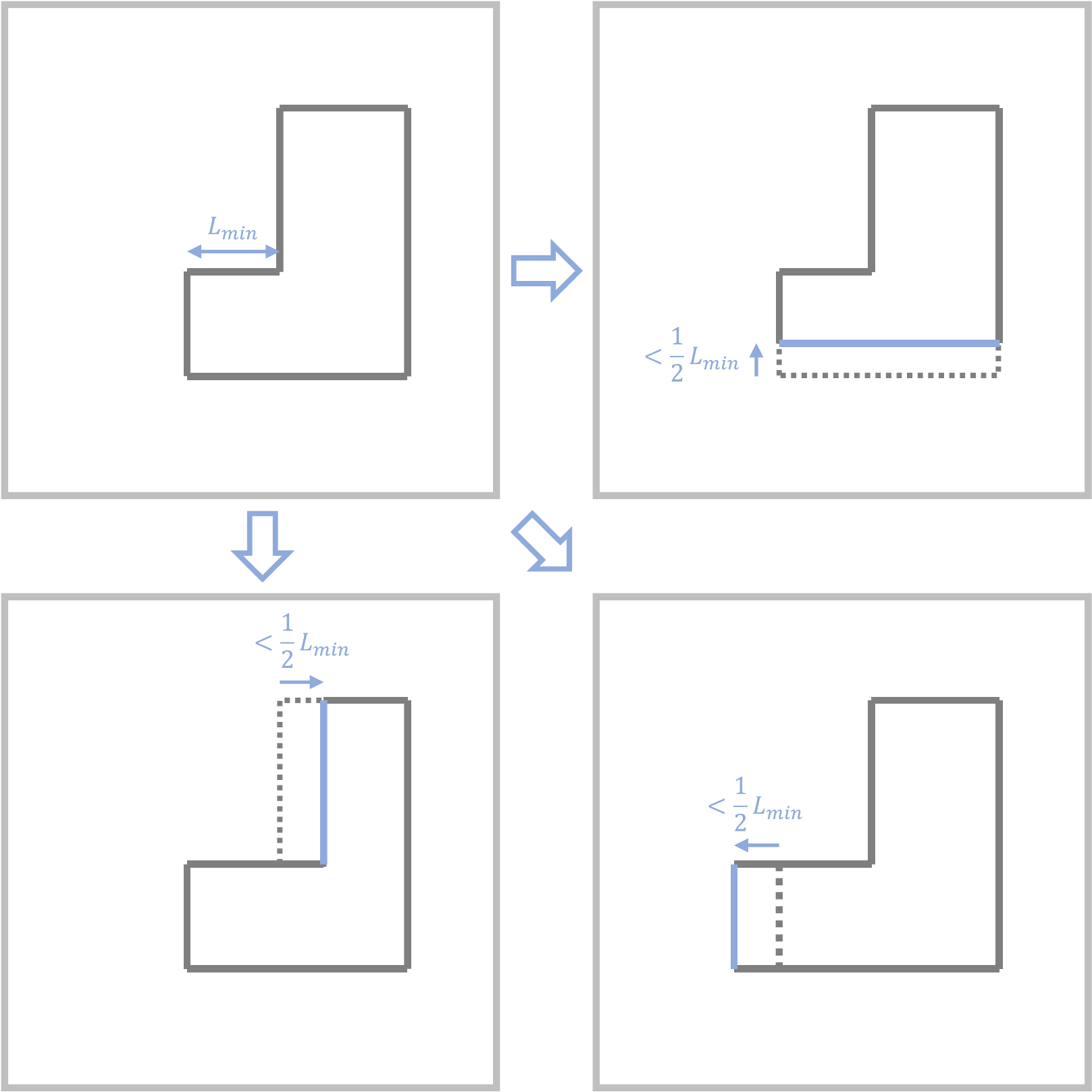}
	\centering
	\caption{
		Conditional uniform data augmentation. 
	}
	\label{fig:aug}
\end{figure}

Finally, we describe a conditional uniform data augmentation strategy. As a reminder, this implicit encoding network is trained between layout occupancy images and functions. This is a domain in which we can easily generate synthetic samples without rendering artifacts. 

\begin{figure*}[ht]
	\includegraphics[width=0.9\linewidth]{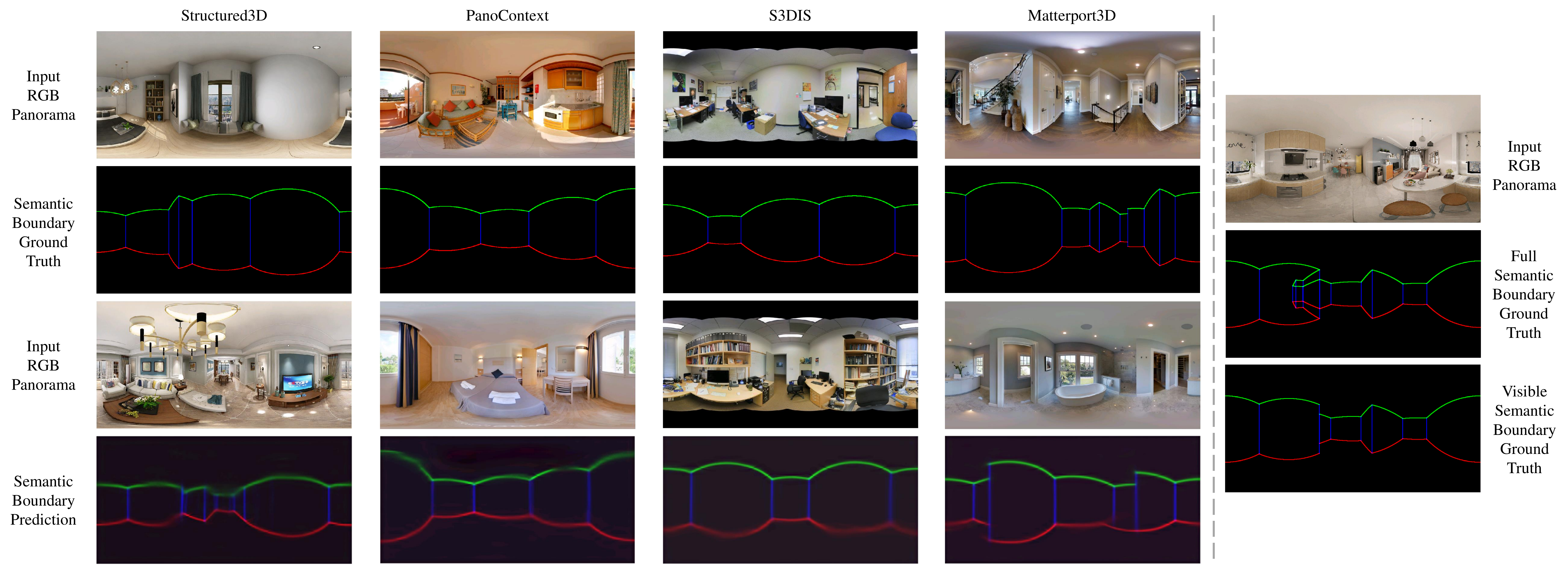}
	\centering
	\caption{
		Left: Semantic boundary map ground truth and prediction. Right: Full and visible ground truth semantic boundary maps.
	}
	\label{fig:fcn}
\end{figure*}

Since training generative models for structured data is still an open problem, we resort to a conditional formulation. As demonstrated in Fig.~\ref{fig:aug}, for every sample in the Structured3D training set, we randomly generate synthetic rooms from it. The first step is to traverse all walls and identify the shortest one with $L_{min}$ length. The second step is to randomly select a wall to augment. The third step is to sample a value from the uniform distribution $[-\frac{1}{2}L_{min},\frac{1}{2}L_{min}]$ and translate the selected wall along its normal according to the value. Augmentation on existing rooms and restricting the translation range guarantee that the synthetic rooms have a reasonable topology. Using this strategy, we generate a total of one million synthetic rooms (referred to as \emph{RoomAug-1M}) to train the implicit encoding network.

\subsection{Shape Code Regression via Image Encoding}

Now the ground truth layout occupancy image is transformed into a latent code by $\pi_\phi^{\rm SE}$. Then we learn an image encoding network to bridge the gap between the input panoramic image and the latent code:

\begin{equation}
	\label{eq:ie}
	\pi_\omega^{\rm IE}:\mathbb{R}^{H\times W\times 3} \rightarrow \mathbb{R}^{D}
\end{equation}

We use a convolutional network to implement $\pi_\omega^{\rm IE}$. We impose $\pi_\phi^{\rm SE}$ on all training samples in datasets to get \emph{(pseudo) ground truth} latent codes. Then we optimize $\omega$ to minimize the $L_2$ or $L_1$ distances between the output of $\pi_\omega^{\rm IE}$ and ground truth latent codes. Formally, we optimize:

\begin{equation}
	\label{eq:lse}
	\arg\min_{\omega}\sum_{i}\frac{1}{2}\|\pi_\omega^{\rm IE}(I_i)-\pi_\phi^{\rm SE}(O_i)\|^2_2
\end{equation}


As such, during test time, we can obtain a set of hyper-planes from a single panoramic image $I_i$ by:

\begin{equation}
	\label{eq:forward}
	P_i=\pi_\psi^{\rm HG}(\pi_\omega^{\rm IE}(I_i))
\end{equation}



As will be shown later in the experiments, $\pi_\omega^{\rm IE}$ shows poor transferability and we successfully address the issue using a data pre-processing step.

\subsection{Data Pre-processing}

We propose a data pre-processing step to map input RGB panoramas into an intermediate domain that is hardly influenced by low-level visual differences and high-level scene composition drift. We choose semantic boundary maps as the intermediate domain (Fig.~\ref{fig:fcn}). 


The original input is a panoramic image $I_i \in \mathbb{R}^{H\times W\times 3}$. The predicted semantic boundary map is denoted by $M_i \in \mathbb{R}^{H\times W\times 3}$. The ground truth semantic boundary map is depicted by $\overline{M}_i \in \mathbb{R}^{H\times W\times 3}$. Three channels of $M_i$ and $\overline{M}_i$ correspond to wall-floor boundary, wall-ceiling boundary and wall-wall boundary respectively. As such the mapping from $I_i$ to $M_i$ is denoted by $\pi_\theta^{\rm SBM}$:

\begin{equation}
	\label{eq:sbn}
	\pi_\theta^{\rm SBM}:\mathbb{R}^{H\times W\times 3} \rightarrow \mathbb{R}^{H\times W\times 3}
\end{equation}

$\pi_\theta^{\rm SBM}$ is a fully convolutional network. As shown in the right panel of Fig.~\ref{fig:fcn}, using visible\footnote{Strictly speaking, here \emph{visible} means being free of layout self-occlusion. It does not mean the map is free of furniture occlusion.} semantic boundary ground truth map is critical. We demonstrate $\{I_i,\overline{M}_i\}$ pairs in the upper rows of Fig.~\ref{fig:fcn}'s left panel. We optimize a pixel-wise $L_2$ loss function between $M_i=\pi_\theta^{\rm SBM}(I_i)$ and $\overline{M}_i$. This data pre-processing step is formally stated as:

\begin{equation}
	\label{eq:lsbn}
	\arg\min_{\theta}\sum_{i}\frac{1}{2}\sum_{i}\|\pi_\theta^{\rm SBM}(I_i)-\overline{M}_i\|^2_2
\end{equation}

Then we apply $\pi_\theta^{\rm SBM}$ on both train and test sets. Semantic boundary prediction maps $M_i$ on unseen test sets are illustrated in the lower two rows of Fig.~\ref{fig:fcn}'s left panel. They are good but not perfect. Complex non-cuboid layout structures in the Structured3D and Matterport3D samples are successfully captured. Yet in the simple PanoContext and S3DIS samples, blurred boundaries caused by occlusion and ceiling decoration still exist. We use inferred imperfect $M_i$ as a surrogate for $I_i$, which significantly improves zero-shot transfer performance. 

\section{Experiments}

\begin{table*}
	\centering
	
	\begin{tabular}{ccccccccc}
		\toprule
		Exp & Encoding & Size & Manhattan & SBM & Loss & RoomAug-1M & IoU-IE (\%) & IoU-LE (\%) \\
		\midrule
		A & FC &  128/16 & $\times$  & $\times$ & $L_2$ & $\times$ & 98.32 & 83.02 \\
		B & GAP & 128/16 & $\times$  & $\times$ & $L_2$ & $\times$ & 96.89 & 81.93 \\
		C & GAP & 256/32 & $\times$  & $\times$ & $L_2$ & $\times$ & 97.24 & 81.98 \\
		D & GAP & 128/16 & $\surd$  & $\times$ & $L_2$ & $\times$ & 97.02 & 82.12 \\
		E & GAP & 128/16 & $\times$  & $\surd$ & $L_2$ & $\times$ & 96.89 & 89.99 \\
		F & GAP & 128/16 & $\times$  & $\surd$ & $L_2$ & $\surd$ & \textbf{98.38} & 90.16 \\
		G & GAP & 128/16 & $\times$  & $\surd$ & $L_1$ & $\surd$ & \textbf{98.38} & \textbf{90.34} \\
		\bottomrule
	\end{tabular}
	\caption{These ablation results are evaluated on Structured3D. Encoding means how representations are mapped to the latent code in $\pi_\phi^{SE}$. Size means the numbers of hyper-planes and shape primitives, i.e., $N_p$ and $N_s$ mentioned above. Manhattan means whether hyper-planes are enforced to two orthogonal directions. SBM means whether the data pre-processing step is used or not. Loss means the loss function used for training $\pi_\omega^{IE}$. RoomAug-1M means whether the augmented synthetic dataset is used.}
	\label{tbl:basic}
\end{table*}

\subsection{Dataset and Evaluation Protocol}

\begin{table*}
	\centering
	\begin{tabular}{ccccccccc}
		\toprule
		\textbf{}  &                &              & \multicolumn{2}{c}{PanoContext} & \multicolumn{2}{c}{S3DIS} & \multicolumn{2}{c}{Matterport3D}                                  \\
		Ft IE & Ft SR & SBM & IoU-IE (\%)    & IoU-LE (\%)    & IoU-IE (\%) & IoU-LE (\%) & \multicolumn{1}{c}{IoU-IE (\%)} & \multicolumn{1}{c}{IoU-LE (\%)} \\
		\midrule
		$\times$ &  $\times$ & $\times$  & 95.94 & 67.04 & 97.35 & 51.43 & \textbf{90.18} & 24.68\\
		$\times$ &  $\times$ & $\surd$  & 95.94 & \textbf{84.43} & 97.35 & \textbf{82.69} & \textbf{90.18} & \textbf{73.73}\\
		\midrule
		$\times$ &  $\surd$ & $\times$  & 95.94 & 51.90 & 97.35 & 44.69 & \textbf{90.18} & 22.75\\
		$\times$ &  $\surd$ & $\surd$  & 95.94 & 81.22 & 97.35 & 80.76 & \textbf{90.18} & 73.26\\
		\midrule
		$\surd$ &  $\surd$ & $\times$  & \textbf{97.05} & 51.08 & \textbf{98.72} & 42.93 & 87.59 & 25.09\\
		$\surd$ &  $\surd$ & $\surd$  & \textbf{97.05} & 81.95 & \textbf{98.72} & 80.25 & 87.59 & 71.06\\
		\bottomrule
	\end{tabular}
	\caption{Transferability evaluation. Ft IE/SR means whether to fine-tune the implicit encoding and shape code regressor, respectively.}
	\label{transfer}
	
\end{table*}

We use four datasets for evaluation: Structured3D \cite{zheng2020structured3d}, PanoContext \cite{zhang2014panocontext}, S3DIS \cite{armeni2016s3dis} and Matterport3D \cite{zou2021manhattan}. Structured3D is a synthetic dataset which can be used to generate diverse annotations. Here we use the version published in the ECCV2020 Holistic 3D workshop, which has 21727 samples. In the alphabetical order of file names, we take one for evaluation every 70 samples. We have 21329 panoramas for training and 308 panoramas for testing. To be consistent with the literature \cite{zou2021manhattan}, we combine the training sets of PanoContext and S3DIS, forming a set of 896 panoramas. For testing, we use the original split, which contains 53 samples for PanoContext and 113 samples for S3DIS. Matterport is also a dataset that allows various usages. We use the layout version annotated by \cite{zou2021manhattan}, which has 1835 panoramas for training and 458 panoramas for testing.

We use 2D intersection over union (IoU) in the top-down viewpoint as our metric. As shown in Fig.~\ref{fig:encoding}, we generate a discretized output (i) and compare it with (a). Pixel-wise intersection and union between (a) and (i) are calculated and we divide intersection by union to get an IoU value. This is used to evaluate the implicit encoding network and referred to as IoU-IE. Meanwhile, we can evaluate this IoU value for end-to-end room layout estimation, which is named as IoU-LE. Besides, we also give a systematic evaluation on the quality of semantic boundary networks. Following former edge detection papers \cite{arbelaez2010contour}, we use the F1-score under optimal dataset scale (ODS) and optimal image scale (OIS) to evaluate the accuracy of semantic boundary prediction. Since the OIS measure allows us to select an optimal scale value for each image, it is always higher than ODS. Three boundary maps are evaluated separately.

\begin{table*}[]
	\centering
	\begin{tabular}{cccccccccc}
		\toprule
		\textbf{}       &                &           &   & \multicolumn{2}{c}{PanoContext} & \multicolumn{2}{c}{S3DIS} & \multicolumn{2}{c}{Matterport3D}                                  \\
		Ft IE & Ft SR & Aug & Loss    & IoU-IE (\%) & IoU-LE (\%)    & IoU-IE (\%) & IoU-LE (\%) & \multicolumn{1}{c}{IoU-IE (\%)} & \multicolumn{1}{c}{IoU-LE (\%)} \\
		\midrule
		$\times$ &  $\times$ & $\times$ & $L_2$  & 95.94 & \textbf{84.43} & 97.35 & 82.69 & 90.18 & 73.73\\
		$\times$ &  $\times$ & $\surd$ & $L_2$  & \textbf{98.91} & 83.60 & \textbf{98.88} & \textbf{82.89} & \textbf{94.79} & \textbf{73.89}\\
		$\times$ &  $\surd$ & $\surd$ & $L_2$  & \textbf{98.91} & 81.48 & \textbf{98.88} & 80.75 & \textbf{94.79} & 72.84\\
		$\times$ &  $\surd$ & $\surd$ & $L_1$  & \textbf{98.91} & 83.30 & \textbf{98.88} & 81.23 & \textbf{94.79} & 73.86\\
		\bottomrule
	\end{tabular}
	\caption{Ft IE/SR means whether to fine-tune the implicit encoding and shape code regression network, respectively. Aug means whether to train the implicit encoding network with RoomAug-1M. Loss means how to train the shape code regressor.}
	\label{aug1m}
\end{table*}

\subsection{Ablation Studies}

The first set of experiments are presented on Structured3D as it is the largest one among four datasets we inspected. We show the impact of basic building blocks in the newly proposed learning system, in Table.~\ref{tbl:basic}. Experiment indexes are shown on the leftmost column. To make comparisons clearer, we use a v.s. operator between indexes.

\textbf{Encoding (A v.s. B):} We study two alternatives for the final encoding step in the mapping $\pi_\phi^{SE}$. The first is to directly use a fully connected layer to map (flattened) convolutional features to the latent shape code. This preserves spatial information which is naturally favorable, since the shape code should have the capability of reconstructing the layout. The second is to apply a global average pooling layer before the fully connected layer. This is an established choice in image recognition \cite{zhou2016learning}, yet whether it is reasonable for implicit encoding still needs justification. Not surprisingly, A outperforms B by 1.43\% for IoU-IE and 1.09\% for IoU-LE. However, this comes at a high cost of parameter usage. As for the implicit encoding network, model A has a size of 33MB while model B has a size of 21MB. As such, we choose the GAP version for all later experiments. 

\textbf{Size (B v.s. C):} We answer the question of how many hyper-planes ($N_p$) and shape primitives ($N_s$) are enough for implicitly encoding generic room layouts. We tried three settings: 64/8, 128/16 and 256/32. The first setting does not converge, implying that it is not enough to approximate the shape space. C outperforms B only by 0.35\% for IoU-IE and 0.05\% for IoU-LE, from which we reach the conclusion that there is no need to further pursue bigger models. 

\textbf{Manhattan (B v.s. D):} The Manhattan constraint is widely used in room layout estimation, thus we explore the possibility of enforcing hyper-plane equations $P_i$ to align with two orthogonal Manhattan directions. Interestingly, D outperforms B by 0.13\% for IoU-IE and 0.19\% for IoU-LE, implying the advantages of incorporating this inductive bias. However, to keep the formulation generic, we do not use the constraint in later experiments.

\textbf{SBM (B v.s. E):} Then we show the impact of obtaining semantic boundary prediction maps as a data pre-processing step. Note that IoU-IE remains the same and E outperforms B by 8.06\% for IoU-LE. This significant margin illustrates the importance of using an intermediate domain in our newly proposed learning system. The mapping $\pi_\phi^{SE}$ alleviates the negative influence of appearance change and provides consistent patterns to recognize. We keep this setting in later experiments due to its effectiveness.

\textbf{RoomAug-1M (E v.s. F):} We then inspect the setting of training with one million augmented rooms. F outperforms E by 1.49\% for IoU-IE and 0.17\% for IoU-LE. As expected, using numerous augmented data better model the shape code space, leading to a clear margin for IoU-IE. As a reminder, RoomAug-1M is generated on Structured3D, thus this margin will get larger on other datasets. This improvement does not translate to IoU-LE, suggesting that other factors like the capability of $\pi_\omega^{IE}$ are the bottleneck.

\textbf{Loss (F v.s. G):} Finally, we use $L_1$ for training $\pi_\omega^{IE}$. Since $L_1$ naturally leads to sparse non-zero entries, if the latent shape code is well-disentangled, using $L_1$ may lead to better performance. Empirically, G outperforms F by 0.18\% for IoU-LE, which is quite marginal.

\subsection{On Transferablility}

In Table.~\ref{transfer}, we show quantitative results on model transferability for PanoContext, S3DIS and Matterport3D. There are three combinations: (1) Directly evaluating on other datasets using models trained on Structured3D, without fine-tuning the implicit encoding nor the shape code regression network on them. This is amounts to a \emph{zero-shot transfer} setting. (2) Only fine-tuning the shape code regression network on other datasets. This setting re-uses shape embeddings trained on Structured3D. (3) Fine-tuning the whole system on other datasets.

\textbf{Poor Transferability of Direct Image Encoding:} The first noticeable fact is that the whole system trained on Structured3D transfers poorly to other datasets, without the data pre-processing step. This is supported by the first row of Table.~\ref{transfer}. The zero-shot transfer results for IoU-IE are 95.94\%, 97.25\% and 90.18\%, respectively. This is understandable as occupancy images/functions are rarely influenced by domain drift and Structured3D is the largest dataset that covers a wide range of room shapes. However, the zero-shot transfer results for IoU-LE are only 67.04\%, 51.43\%, and 24.68\%, respectively. This shows that direct image encoding suffers from severe domain drift. 

\begin{table}
	\centering
	
	\begin{tabular}{cccc}
		\toprule
		\textbf{S3D} & Floor & Ceiling & Wall\\
		\midrule
		Train &  63.53/60.70 & 63.97/54.27  & 68.79/63.82\\
		Test &  63.68/61.12 & 64.34/54.42  & 69.16/63.60\\
		All &  63.53/60.71 & 63.95/54.28 & 68.80/63.82\\
		
		\midrule
		\textbf{P\&S} & Floor & Ceiling & Wall\\
		\midrule
		Train &  50.13/47.41 & 52.49/49.53  & 52.80/48.99\\
		Test-P &  46.02/44.09 & 42.92/39.14  & 42.78/37.79\\
		Test-S &  44.96/42.71 & 56.07/53.84  & 45.80/40.94\\
		All &  49.27/46.67 & 52.39/49.35 & 51.56/47.53\\
		
		\midrule
		\textbf{MP3D} & Floor & Ceiling & Wall\\
		\midrule
		Train &  45.85/43.44 & 44.75/41.11  & 46.35/41.46\\
		Test &  44.75/42.18 & 38.97/35.27  & 44.18/38.99\\
		All &  45.63/43.17 & 43.60/39.85 & 45.92/40.95\\
		\bottomrule
		
	\end{tabular}
	\caption{Semantic boundary prediction accuracy evaluation. Numbers before and after the slash are F1-score under OIS and ODS respectively, which are measured in \%.}
	\label{tbl:s3dse}
\end{table}

\textbf{Semantic Boundary Maps Help:} As demonstrated in the second row of Table.~\ref{transfer}, introducing the intermediate domain significantly improves the zero-shot transfer performance, which the most important finding of this paper. Note that IoU-IE is not influenced and IoU-LE sees a performance boost of 17.39\%, 31.26\% and 49.05\%, respectively. This is achieved without fine-tuning the latent space or the shape code regressor on other datasets. As a conclusion, the usage of intermediate domain largely improves the generalization ability of our method, because the domain drift of biased shape code regression is effectively addressed by bypassing the negative influences of low-level appearance change and high-level scene composition change. 

\textbf{Does Fine-tuning the Shape Code Regressor Help?} Very interestingly, the answer is no, as evidenced by the third and fourth row of Table.~\ref{transfer}. Without SBM, fine-tuning the shape code regressor leads to 15.14\%, 6.74\% and 1.93\% performance drops, respectively.  With SBM, it leads to 3.21\%, 1.93\% and 0.47\% performance drops, respectively. These results conclusively show that fine-tuning the shape code regressor on small datasets results in over-fitting and severely hurts model performance. This fact further confirms the significance of SBM because the most straightforward solution of fine-tuning $\pi_\omega^{IE}$ on target datasets cannot address the domain drift of biased shape code regression. 

\textbf{Does Fine-tuning the Implicit Encoding Network Help?} The answer is still no, as evidenced by the last two rows of Table.~\ref{transfer}. Without SBM, fine-tuning the whole system leads to -0.82\%, -1.76\% and +2.34\% performance changes, respectively. With SBM, it leads to +0.73\%, -0.51\% and -2.20\% performance changes, respectively. The margins do not point to clear conclusions but it is demonstrated that even fine-tuning the whole system on target datasets cannot fully address the domain drift problems.

\subsection{Impact of RoomAug1M}

Table.~\ref{aug1m} summarizes the quantitative results on training the implicit encoding network with RoomAug1M. As demonstrated by the second row, introducing RoomAug1M significantly promotes IoU-IE on three small datasets in the zero-shot transfer setting. We achieve 2.97\%, 1.53\% and 
4.61\% boosts on PanoContext, S3DIS and Matterport3D, respectively. Notably, IoU-IE values on PanoContext and S3DIS are as high as 98.91\% and 98.88\%, which are nearly perfect. This validates our assumption that using synthetic data in a simple domain free of rendering artifacts is a good practice. Meanwhile, the IoU-LE values only see little performance drop or increase, implying that other factors are becoming bottlenecks that prevent us from fully unleashing the power of a better shape code space. 

\textbf{Fine-tuning the Shape Code Regressor Still Hurts Performance:} With the RoomAug1M dataset used, fine-tuning the implicit encoding network on small datasets become meaningless. But we still investigate the option of fine-tuning the shape code regressor. As shown in the third row of Table.~\ref{aug1m}, this stills brings 2.12\%, 2.14\% and 1.05\% performance drops, respectively. This indicates that there is two sources of domain drift. Even when the domain drift of a biased shape code space is alleviated by enormous data, fine-tuning the shape code regressor on small datasets still amplifies the domain drift of a biased shape code regressor.

\textbf{$L_1$ Regression Empirically Helps:} As shown in the last row of Table.~\ref{aug1m}, switching to a $L_1$ loss function for shape code regression brings 1.82\%, 0.48\%, and 1.02\% improvements, respectively, although the results are still lower than the zero-shot transfer setting. This suggests the shape code space learned on RoomAug1M may have some disentanglement characteristics. 

\subsection{Semantic Boundary Accuracy}

Lastly, we provide quantitative evaluations for semantic boundary map quality, in Table.~\ref{tbl:s3dse}. S3D means Structured3D, P\&S means PanoContext and S3DIS, and MP3D means Matterport3D. As a reminder, the training sets of PanoContext and S3DIS are combined. Since $\pi_\theta^{SBM}$ is used as a data pre-processing step, we impose it on both training and testing images. The semantic boundary map prediction accuracy on Structure3D is the highest, which directly translates to its high IoU-LE (90.34\%). 

\section{Conclusion}

We propose an end-to-end room layout estimation method taking panoramic images as inputs. It is based upon the implicit encoding principle. We first learn shape codes that serve as a surrogate for a structured implicit representation. The representation is self-supervised by a combination of several differentiable rendering layers. The we learn a mapping from input images to the latent shape code, which makes end-to-end room layout estimation possible. Like other end-to-end formulations, the proposed one is troubled by generalization ability. On one hand, we show a data pre-processing step significantly improves the zero-shot transfer performance. One the other hand, we propose a conditional uniform data augmentation strategy that alleviates the domain drift of a biased shape code space. Extensive evaluations are conducted on four widely used public datasets, with code, data and models released.

\textbf{Limitation:} The current representation exploits a fixed number of hyper-planes, which is not compatible with several widely used metrics on those four benchmarks.

{\small
\bibliographystyle{ieee_fullname}
\bibliography{egbib}
}

\begin{figure*}[ht]
	\includegraphics[width=0.8\linewidth]{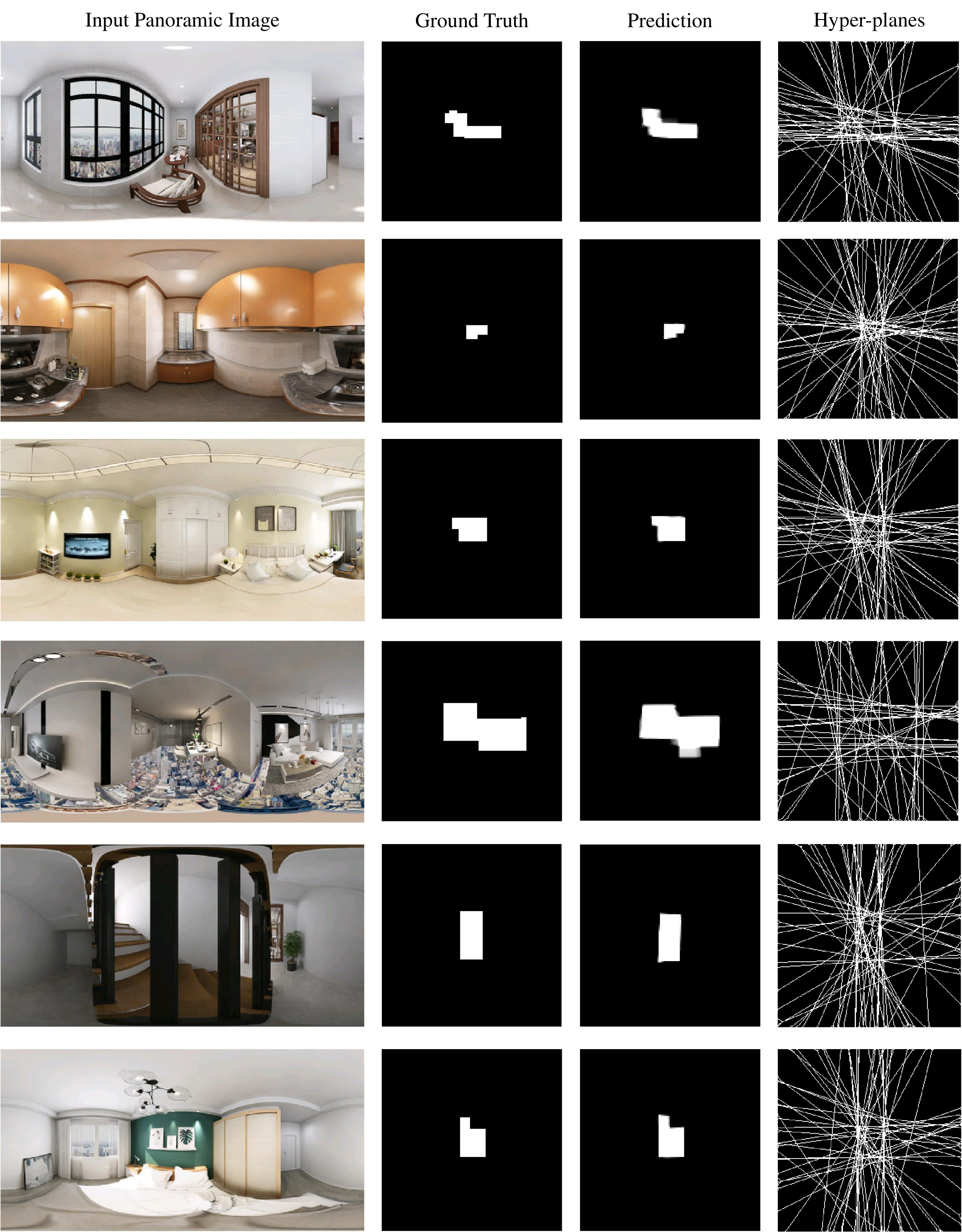}
	\centering
	\caption{
		Qualitative results on Structured3D. SBM and RoomAug1M are used.
	}
	\label{fig:s3dquali}
\end{figure*}

\begin{figure*}[ht]
	\includegraphics[width=0.8\linewidth]{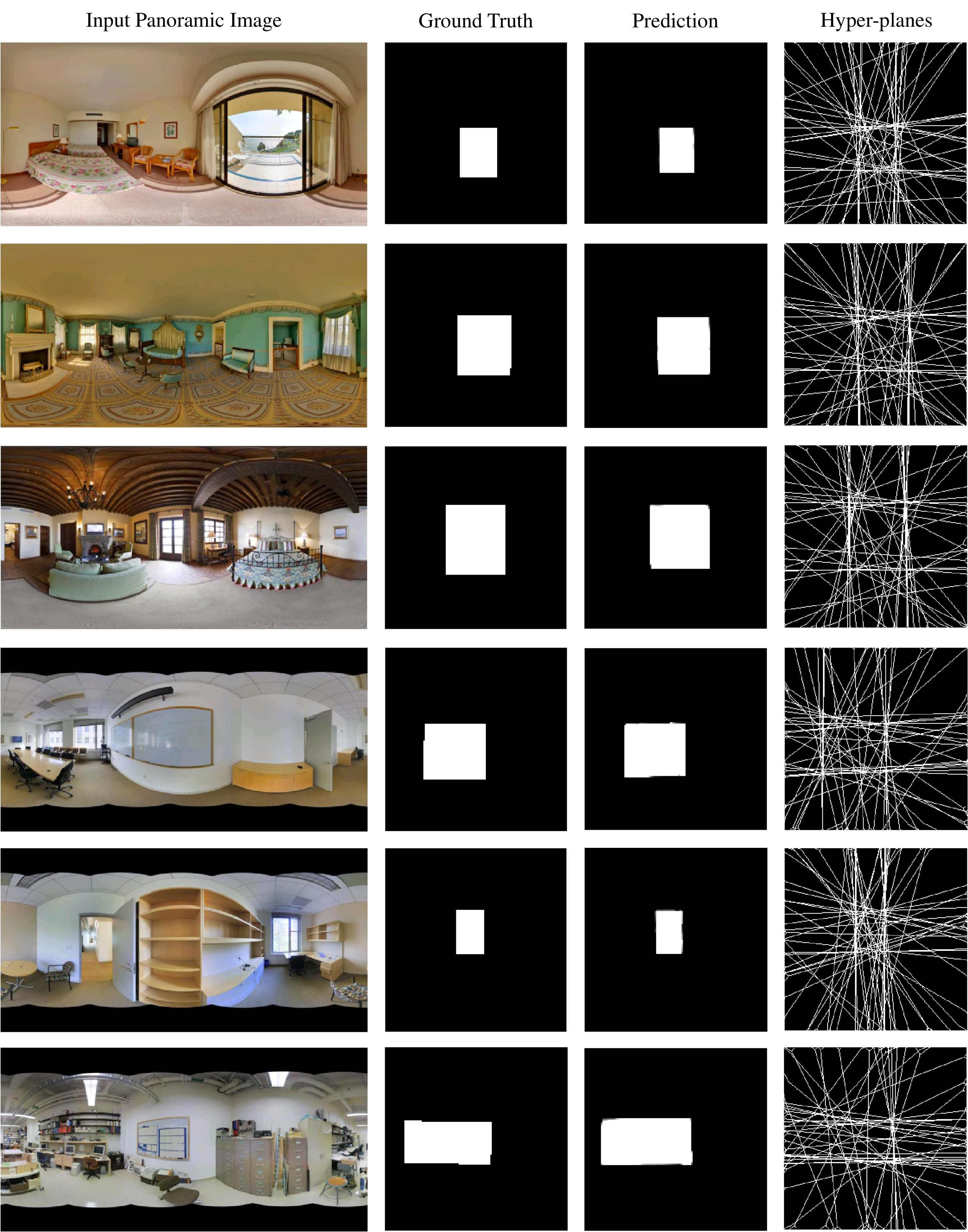}
	\centering
	\caption{
		Qualitative results on PanoContext (upper three) and S3DIS (lower three), in the zero-shot transfer setting. SBM and RoomAug1M are used.
	}
	\label{fig:lnquali}
\end{figure*}

\begin{figure*}[ht]
	\includegraphics[width=0.8\linewidth]{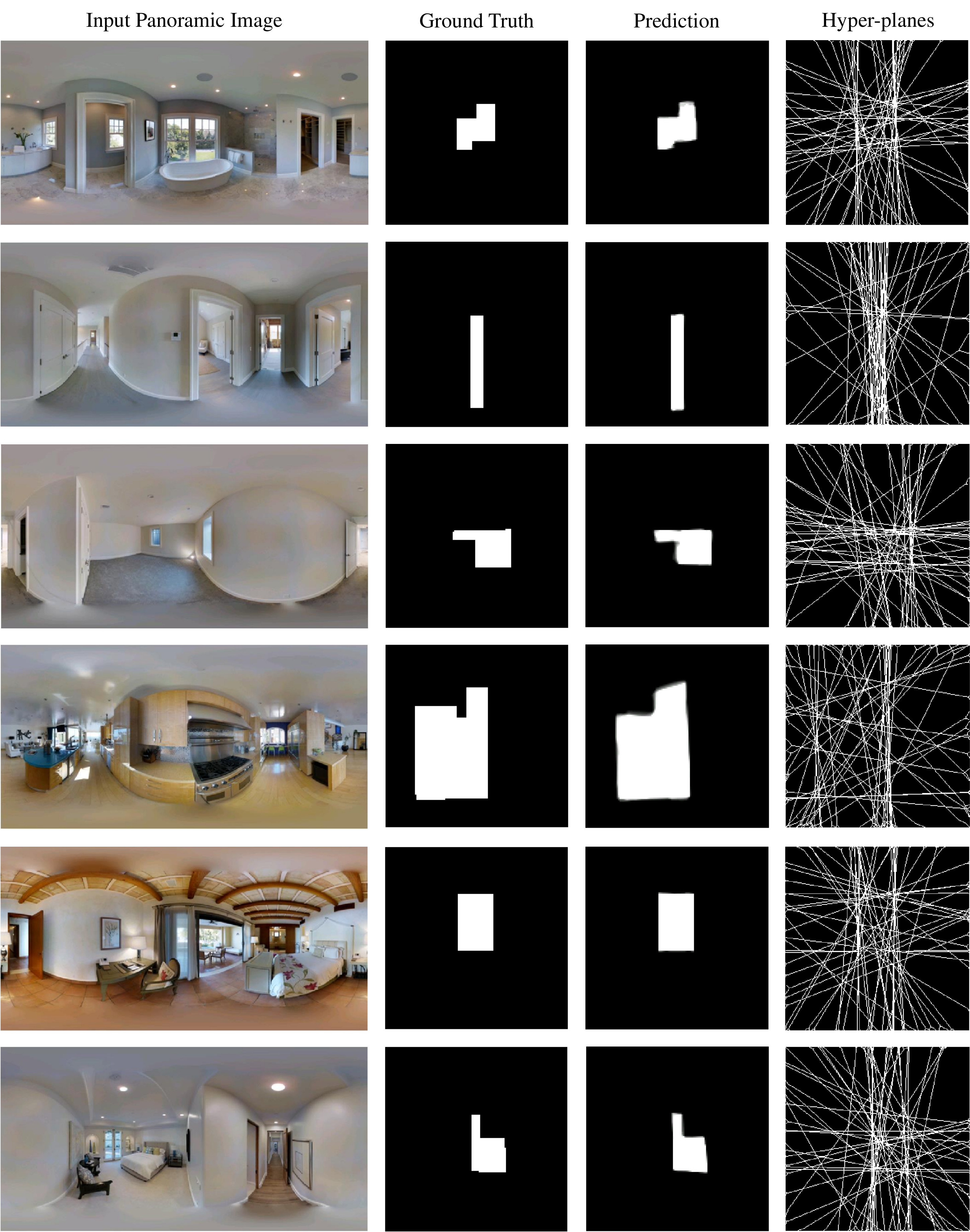}
	\centering
	\caption{
		Qualitative results on Matterport3D, in the zero-shot transfer setting. SBM and RoomAug1M are used.
	}
	\label{fig:mp3dquali}
\end{figure*}

\end{document}